\documentclass[letterpaper,journal]{IEEEtran}
\usepackage{amsmath,amsfonts}
\usepackage{amssymb,bm}
\usepackage{algorithmic}
\usepackage{algorithm}
\usepackage{array}
\usepackage{booktabs,makecell}
\usepackage{multirow}
\usepackage[caption=false,font=normalsize,labelfont=sf,textfont=sf]{subfig}
\usepackage{textcomp}
\usepackage{stfloats}
\usepackage{url}
\usepackage{verbatim}
\usepackage{graphicx}
\usepackage{cite}
\begin{document}

\title{LCMamNet: A Lightweight Cross-scale Mamba Network for Infrared Small Target Detection}

\author{Yuhao~Fan,
	Le~Hui,
	and~Yuchao~Dai%
	\thanks{(Corresponding authors: Le Hui; Yuchao Dai.)}%
	\thanks{Yuhao Fan, Le Hui, and Yuchao Dai are with the School of Electronics
		and Information, Northwestern Polytechnical University, Xi'an 710072, China
		(e-mail: fanyuhao@mail.nwpu.edu.cn; huile@nwpu.edu.cn;
		daiyuchao@nwpu.edu.cn).}}

\maketitle
\begin{abstract}
	Infrared small target detection (IRSTD) is important for low-altitude perception, unmanned-system warning, and security monitoring. However, weak targets in infrared imagery usually occupy only a few pixels and are easily submerged by cloud clutter, ground edges, and bright noise, making it difficult for lightweight segmentation-based methods to preserve local target structures while suppressing background interference. To address these challenges, we propose LCMamNet, a lightweight cross-scale Mamba network that progressively enhances local target structures, interacts cross-scale context in a latent space, and restores spatial details with background suppression. Specifically, a compact hierarchical encoder with cross-shaped directional bottleneck residual (CDBR) blocks strengthens direction-sensitive target structures under a small computation budget. A latent dense cross-scale fusion (LDCF) module then performs dense all-level interaction through bidirectional Mamba modeling and reorganizes the interacted features into stable hierarchical semantics. Finally, a progressive decoder selectively recovers shallow spatial details while suppressing irrelevant background textures. Extensive experiments on IRSTD-1k, NUAA-SIRST, and NUDT-SIRST show that the proposed network achieves mIoU scores of 71.25\%, 79.60\%, and 95.58\%, respectively, with only 1.175M parameters and 6.91 GFLOPs. It also runs with a mean inference latency of 6.62 ms, and deployment results on an NVIDIA Jetson Orin NX 16G SUPER further demonstrate its practical potential for real-time edge inference. The code and checkpoints are publicly available at \url{https://github.com/Haoyu096/LCMamNet}.
\end{abstract}

\begin{IEEEkeywords}
infrared small target detection, lightweight network, Mamba, cross-scale fusion, edge deployment
\end{IEEEkeywords}

\section{Introduction}
\IEEEPARstart{I}{nfrared} small target detection (IRSTD) aims to segment tiny targets from infrared images and plays an important role in low-altitude perception, unmanned-system warning, security monitoring, and target reconnaissance\cite{sf_irstd_survey,rethink_generalizable_irstd,rethink_eval_irstd}. Unlike common objects in natural images, infrared small targets usually occupy only a handful of pixels and exhibit weak responses, low contrast, blurred boundaries, and scarce texture. Meanwhile, cloud structures, ground textures, strong edges, and bright noise can easily overwhelm target responses and lead to false alarms. Therefore, practical IRSTD requires a detector that can preserve weak target structures, distinguish them from cluttered backgrounds, and remain efficient enough for resource-constrained deployment.

\begin{figure}[!t]
	\centering
	\includegraphics[width=0.95\columnwidth]{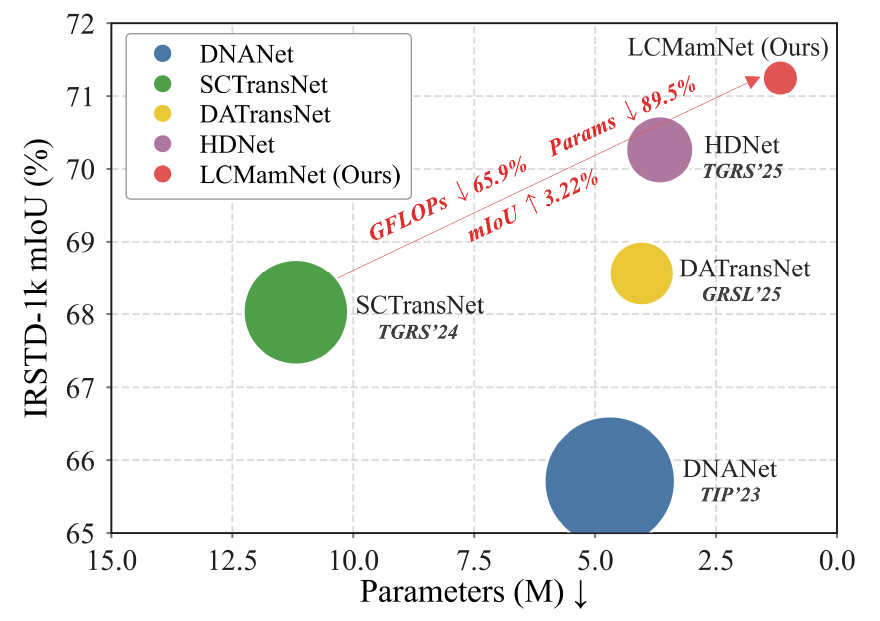}
	\caption{Accuracy--efficiency trade-off among representative deep infrared small target detection models on IRSTD-1k. The horizontal axis denotes parameter count in reversed order, the vertical axis denotes mIoU, and the area of each circle is proportional to GFLOPs. Therefore, a point closer to the upper-right corner indicates higher mIoU with fewer parameters. The red arrow marks the shift from SCTransNet to our method, which achieves higher mIoU together with lower parameters and GFLOPs.}
	\label{fig:tradeoff_teaser}
\end{figure}

With the development of deep learning, segmentation-based methods have become the dominant solution for single-frame IRSTD. CNN-based U-shaped models\cite{unet} reduce target loss during downsampling through local contrast enhancement, multi-scale aggregation, and cross-level feature reuse\cite{acm,dnanet,alcnet,uiunet}. However, their local receptive fields still limit contextual discrimination under complex backgrounds. Attention-based methods strengthen semantic discrimination through cross-level interaction\cite{sctransnet,datransnet}, while existing Mamba-based IRSTD methods mainly embed state-space modeling into backbone or block-level feature extraction for efficient long-range dependency representation\cite{mim_istd,hmcnet}. These designs improve semantic discrimination, but they often increase architectural complexity and may disturb scale-specific semantic organization after repeated multi-level interaction. As illustrated in Fig.~\ref{fig:tradeoff_teaser}, a favorable balance between segmentation accuracy and model complexity remains difficult to achieve.

For segmentation-based IRSTD, the above trade-off is closely related to three coupled requirements. First, weak targets must be protected in early encoding stages, since attenuated local structures and directional cues are difficult to recover by later context modeling alone. Second, cross-scale interaction should enrich contextual dependency without blurring the scale attribution and semantic organization of different feature levels. Third, shallow details should be restored selectively, because skip features contain useful target boundaries as well as substantial background texture. These requirements suggest that lightweight IRSTD should coordinate local structure enhancement, cross-scale contextual modeling, and selective detail restoration, instead of relying on a single stronger interaction module.

To address these issues, we propose LCMamNet, a lightweight infrared small target detection network with a coordinated encoder-fusion-decoder design. First, a hierarchical encoder built with cross-shaped directional bottleneck residual (CDBR) blocks enhances direction-sensitive local target structures under a compact computation budget. Second, a latent dense cross-scale fusion (LDCF) module formulates deeper multi-level feature fusion as latent-space interaction, where bidirectional Mamba modeling\cite{mamba,visionmamba,vmamba} establishes dense cross-scale association and hierarchy-aware semantic reorganization restores stable multi-level semantics. Finally, a progressive decoder selectively restores shallow spatial details while suppressing background texture leakage. Through this flow, the proposed network keeps the benefits of cross-scale context modeling while maintaining lightweight local representation and controlled detail restoration.

Experimental results on three public infrared small target detection benchmarks show that LCMamNet achieves strong region-level segmentation performance with low complexity. It reaches mIoU scores of 71.25\%, 79.60\%, and 95.58\% on IRSTD-1k~\cite{isnet}, NUAA-SIRST~\cite{acm}, and NUDT-SIRST~\cite{dnanet}, respectively, with only 1.175M parameters and 6.91 GFLOPs. Deployment experiments on a Jetson Orin NX 16G SUPER further verify its potential for real-time edge inference. The main contributions are summarized as follows:
\begin{enumerate}
	\item We propose LCMamNet, a lightweight infrared small target detection network that coordinates local structure encoding, latent cross-scale fusion, and selective detail restoration. Experiments on public benchmarks and an edge deployment platform verify its favorable balance among accuracy, complexity, and inference efficiency.
	
	\item We design a CDBR-based hierarchical encoder that combines cross-shaped directional convolution with bottleneck residual learning to strengthen direction-sensitive weak-target structures at low complexity.
	
		\item We design LDCF as a latent cross-scale interaction path that integrates bidirectional Mamba modeling with hierarchy-aware semantic reorganization for deeper encoder features, while selective progressive decoding restores shallow details and suppresses background interference.
\end{enumerate}

\section{Related Work}
This section reviews representative IRSTD methods from three aspects: CNN-based architectures, context and cross-scale interaction, and Mamba-based modeling. The discussion focuses on a common issue behind these directions: how to enhance weak-target representation while preserving stable scale-specific semantics under a compact computational budget.
\subsection{CNN-Based IRSTD Methods}
Early IRSTD methods mainly relied on handcrafted target-background priors, including directional filtering, local contrast measurement, low-rank sparse decomposition, and tensor-based modeling. Representative methods include the edge directional LMS filter~\cite{edge_lms}, IPI~\cite{ipi}, LCM~\cite{lcm}, MPCM~\cite{mpcm}, low-rank sparse representation~\cite{lowrank_sparse}, and later extensions based on entropy weighting, random walks, and tensor priors~\cite{entropy_weighted,ript,vmd,fkrw,pstnn}. Although these methods offer interpretable assumptions about small-target saliency, they are often sensitive to clutter, low signal-to-noise ratios, and scene-dependent parameter settings.

With deep learning, segmentation-based CNN models have become the mainstream solution for single-frame IRSTD. MDvsFA~\cite{mdvsfa}, ACM~\cite{acm}, ALCNet~\cite{alcnet}, AGPCNet~\cite{agpcnet}, and ISNet~\cite{isnet} improve target-background discrimination through task-specific feature learning, attention, and asymmetric context modeling. Later U-shaped networks further strengthen weak-target preservation with dense interaction, nested decoding, and multi-scale supervision, as shown in DNANet~\cite{dnanet}, ISTDU-Net~\cite{istdunet}, UIU-Net~\cite{uiunet}, MSHNet~\cite{mshnet}, and ILNet~\cite{ilnet}. Recent studies also explore task-specific local operators for weak infrared targets. For example, pinwheel-shaped convolution (PConv) uses multi-directional asymmetric branches to better fit dim small-target distributions with limited parameter overhead~\cite{pconv}. FLINet~\cite{flinet} introduces fuzzy logic to enhance and fuse infrared small-target features, while MSSLNet~\cite{msslnet} adopts multiscale sparse large-kernel modeling to enlarge effective local context. These CNN-based methods provide effective local representation and recovery paths, but their convolutional operators still have limited ability to capture long-range contextual dependencies and global semantic relations in complex infrared scenes.

\subsection{Context Modeling and Cross-Scale Interaction for IRSTD}
Recent IRSTD studies have therefore introduced stronger context modeling and cross-scale interaction mechanisms. These designs often revise skip connections, inter-level fusion paths, or global aggregation modules to improve information exchange among encoder and decoder stages. SCTransNet~\cite{sctransnet} embeds spatial-channel cross Transformer blocks into long skip connections to promote full-level semantic interaction. DATransNet~\cite{datransnet} uses dynamic attention to adaptively enhance weak-target responses, while HSTNet~\cite{hstnet} introduces sparse spatial-channel Transformer modeling to separate targets from cluttered backgrounds. HDNet~\cite{hdnet} further combines spatial-domain CNN features with frequency-domain enhancement to suppress background interference. More recent works continue to refine feature interaction from different perspectives: PQGNet~\cite{pqgnet} uses perceptual-query guidance, MPCNet~\cite{mpcnet} designs multiscale perception and cross-attention feature fusion, and FGARNet~\cite{fgar} introduces frequency guidance with aliasing rectification for small-target enhancement.

Cross-scale interaction is crucial for IRSTD because small targets require both shallow detail cues and deeper contextual discrimination. MSHNet~\cite{mshnet} and MDS-DWGANet~\cite{mds_dwganet} further explore hierarchical similarity modeling, direction-aware convolution, and dynamic attention for multi-level feature exchange. However, stronger interaction is not always equivalent to better semantic organization. Repeated multi-level mixing may blur scale attribution or disturb spatial detail, especially when weak targets occupy only a few pixels. This motivates a fusion strategy that can build sufficient contextual dependency while explicitly reorganizing scale-aware semantics before decoding.

\subsection{Mamba-Based Modeling for IRSTD}
Mamba, derived from selective state space models, provides an efficient way to model long-range dependencies with input-dependent state transitions and linear-time computation\cite{mamba}. In vision tasks, Vision Mamba~\cite{visionmamba} and VMamba~\cite{vmamba} show that state-space modeling can serve as an alternative to attention-based global modeling. Subsequent studies extend this idea through local scanning, multi-scale hierarchies, hybrid backbones, and segmentation-oriented designs, including LocalMamba~\cite{localmamba}, Multi-Scale VMamba~\cite{msvmamba}, MambaVision~\cite{mambavision}, and U-Mamba~\cite{umamba}.

For IRSTD, MiM-ISTD~\cite{mim_istd} introduces nested Mamba modeling to capture both global and local information, and HMCNet~\cite{hmcnet} combines Mamba with CNN-based encoder-decoder structures for weak-target segmentation. These works indicate that state-space modeling is promising for efficient long-range dependency representation. Nevertheless, IRSTD is highly sensitive to local target structure and scale-specific semantic consistency. Long-range modeling alone cannot guarantee that tiny targets are preserved after downsampling and cross-scale fusion. Therefore, a Mamba-based IRSTD framework still needs to coordinate efficient latent-space interaction with local structure encoding and hierarchy-aware semantic restoration. Guided by this consideration, this work places bidirectional state-space modeling in a later latent cross-scale fusion stage, where deeper multi-level features are projected into a unified sequence for cross-scale association, while local structure encoding and hierarchy-aware semantic reorganization are preserved as separate functions.

\begin{figure*}[!t]
	\centering
	\includegraphics[width=\textwidth]{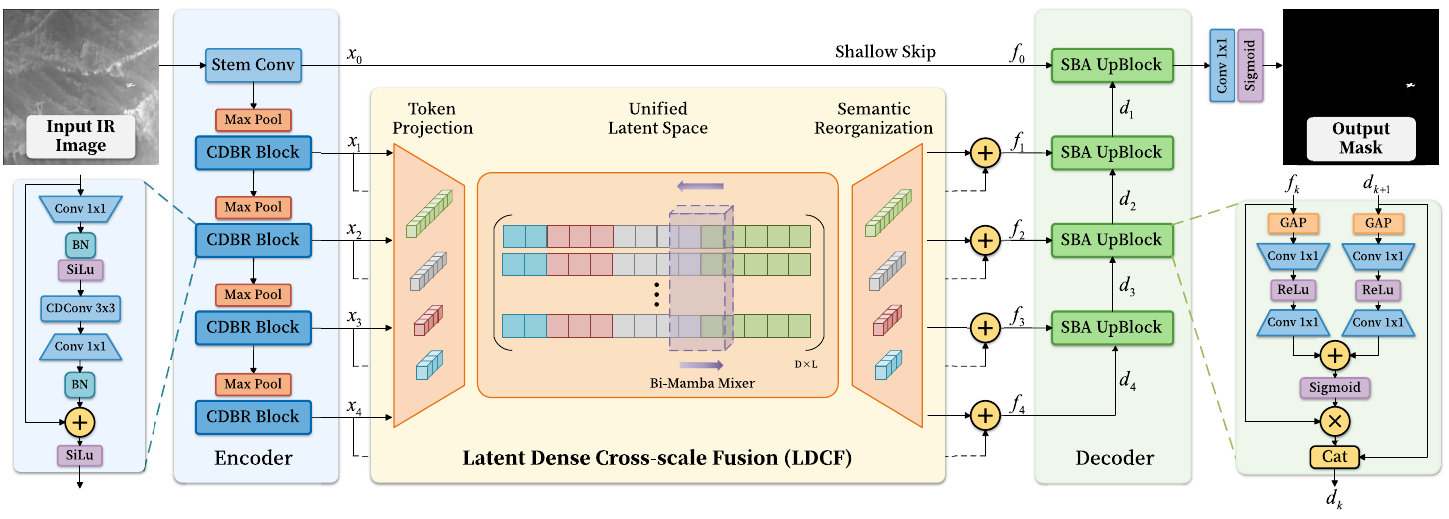}
	\caption{Overall architecture of LCMamNet. Given an input infrared image, the stem layer first extracts the shallow feature $\mathbf{x}_0$, while the Cross-shaped Directional Bottleneck Residual (CDBR)-based hierarchical encoder produces deeper multi-level features $\mathbf{x}_1$ to $\mathbf{x}_4$. These deeper features are then projected into a unified latent space and fused by the Latent Dense Cross-scale Fusion (LDCF) module. Finally, the progressive decoder combines the decoded features with the preserved shallow skip through a lightweight Synergistic Bottleneck Attention (SBA) unit to generate the final prediction mask.}
	\label{fig:overall_architecture}
\end{figure*}

\section{Method}
This section first presents the overall architecture of the proposed network and then details the two main technical components: the hierarchical encoder for local structure preservation and the latent fusion module for cross-scale context interaction. The final decoding process is described in the overall pipeline as a lightweight restoration path for regulated shallow-detail recovery.

\subsection{Overall Architecture}

\subsubsection{Forward Pipeline}

To simultaneously preserve weak-target local structures, model cross-scale context, and suppress shallow noise under a lightweight budget, LCMamNet follows an encoder-fusion-decoder pipeline, as shown in Fig.~\ref{fig:overall_architecture}. Given an input infrared image, the stem layer first extracts the shallow feature $\mathbf{x}_0$, and a cross-shaped directional bottleneck residual (CDBR)-based hierarchical encoder then produces deeper multi-level features. These deeper features are projected into a unified latent space by a latent dense cross-scale fusion (LDCF) module, where cross-scale interaction and hierarchy-aware semantic reorganization are performed before decoding. Finally, a progressive decoder restores spatial details, and a lightweight synergistic bottleneck attention (SBA) unit regulates the preserved shallow skip before prediction.

Given an input infrared image $\mathbf{I}\in\mathbb{R}^{1\times H\times W}$, the network first extracts an initial shallow feature $\mathbf{x}_0$ through a stem layer, and then generates four deeper-scale features along the hierarchical encoding path:
\begin{equation}
	\{\mathbf{x}_i\}_{i=1}^{4}.
\end{equation}
Unlike conventional residual encoders\cite{resnet,mobilenetv2}, the hierarchical encoder is built to maintain weak-target local geometry and direction-sensitive responses during progressive downsampling. The resulting deeper features serve as the semantic input to the subsequent fusion stage, while the shallowest feature is intentionally kept outside the latent interaction path.

\subsubsection{Shallow Feature Routing}

The shallowest feature $\mathbf{x}_0$ is excluded from latent cross-scale fusion and is instead preserved as a shallow-detail anchor for the final restoration stage. This routing choice follows two practical considerations. First, $\mathbf{x}_0$ has the highest spatial resolution, and projecting it into latent tokens would substantially lengthen the unified sequence and weaken the lightweight objective. Second, this feature mainly contains edges, local contrast, and fine-grained spatial responses, which are useful for boundary recovery but may also carry high-frequency background textures. Directly mixing it with deeper semantic features may therefore introduce shallow clutter into the latent interaction process.

Therefore, only $\mathbf{x}_1$--$\mathbf{x}_4$ are used for latent cross-scale fusion, while $\mathbf{x}_0$ is reserved as a shallow skip. After the deeper features are fused and reorganized, the decoder introduces this shallow-detail anchor only near the final restoration stage, where the attention unit regulates skip information before prediction. In this way, the network separates latent contextual interaction from shallow-detail recovery, allowing cross-scale semantics to be formed before high-resolution local details are selectively restored. This design follows the general logic of preserving high-resolution detail branches in encoder-decoder frameworks\cite{unet,fpn}.

As illustrated by the enlarged SBA UpBlock in Fig.~\ref{fig:overall_architecture}, SBA is used as a lightweight skip-regulation unit in the decoder. Given the preserved skip feature $\mathbf{f}_k$ and the upsampled decoder feature $\mathbf{d}_{k+1}$, SBA first extracts channel descriptors from both branches through global average pooling and two lightweight $1\times1$ transformations. The two branch responses are summed and passed through a sigmoid function to generate a channel-wise gate, which recalibrates the skip feature before concatenation with the decoder feature. Different from direct skip concatenation, this design allows decoder semantics to regulate high-resolution shallow details, reducing the transfer of background textures while keeping useful boundary cues for final mask recovery.

\begin{figure*}[!t]
	\centering
	\includegraphics[width=1.0\textwidth]{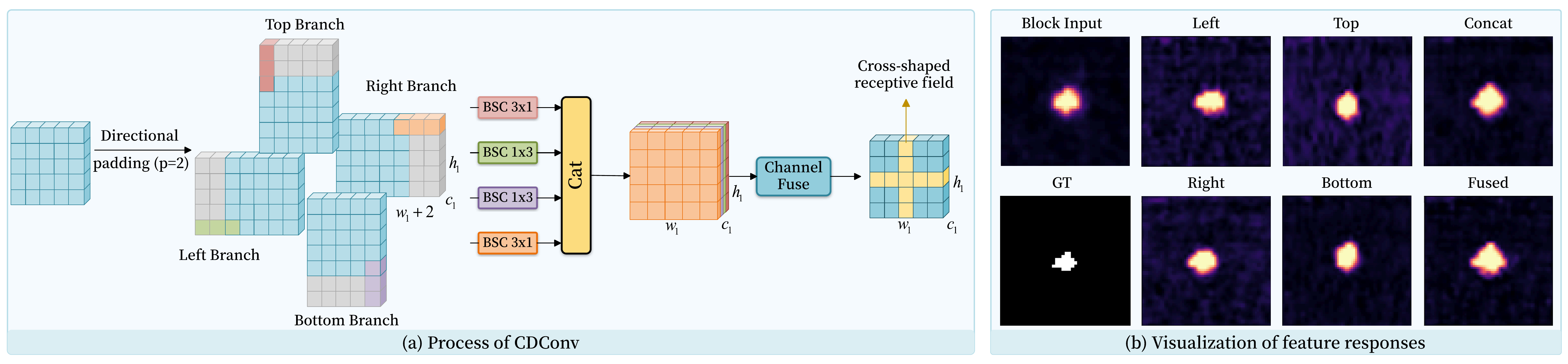}
	\caption{Illustration of the proposed Cross-shaped Directional Convolution (CDConv) and its feature responses. (a) Process of CDConv, where a cross-shaped directional representation is built through directional padding, branch-specific spatial convolutions (BSCs), concatenation (Cat), and Channel Fuse. (b) Visualization of feature responses inside the Cross-shaped Directional Bottleneck Residual (CDBR) block. The locally enlarged maps include the block input, ground truth (GT), the outputs of the four BSC branches, the concatenated branch output, and the channel-fused output.}
	\label{fig:cdconv_cdbr}
\end{figure*}

\subsection{CDBR-Based Hierarchical Encoder}

For infrared small target detection (IRSTD), the goal of the encoding stage is not merely to extract deeper semantic features, but, more importantly, to preserve weak-target local structures and directional cues throughout repeated downsampling. Because infrared small targets occupy very few pixels, their local responses are easily submerged by cloud textures, ground edges, and bright background noise. Once shallow structural information is weakened at early stages, even stronger contextual modeling later on can hardly recover the lost target representation. We therefore introduce a cross-shaped directional bottleneck residual (CDBR) block into the hierarchical encoder to enhance the network's sensitivity to weak-target local structures under lightweight constraints and to maintain more stable feature extraction.

\subsubsection{Cross-Shaped Directional Convolution}

The core local operator in this block is cross-shaped directional convolution (CDConv), which decomposes local spatial modeling into four orthogonal directional branches. Given an input feature $\mathbf{X}\in\mathbb{R}^{C\times H\times W}$, the block first compresses the channels into a lower-dimensional intermediate space through a $1\times1$ convolution:
\begin{equation}
	\mathbf{U}=\mathcal{F}_{\mathrm{red}}(\mathbf{X}), \quad \mathbf{U}\in\mathbb{R}^{C_b\times H\times W}
	\label{eq:cdbr_reduce}
\end{equation}
where $C_b$ denotes the bottleneck channel dimension. CDConv is then applied in this compressed space for direction-sensitive local enhancement. For a kernel size of $k=3$, the asymmetric padding operators for the four directions are defined as
\begin{equation}
	\begin{aligned}
		\mathcal{P}_{\mathrm{l}} &= (2,0,0,0), \quad
		\mathcal{P}_{\mathrm{r}} = (0,2,0,0) \\
		\mathcal{P}_{\mathrm{t}} &= (0,0,2,0), \quad
		\mathcal{P}_{\mathrm{b}} = (0,0,0,2)
	\end{aligned}
	\label{eq:cdconv_padding}
\end{equation}
where each quadruple denotes the zero-padding widths on the left, right, top, and bottom sides, respectively. Based on these definitions, the responses of the four directional branches can be written as
\begin{equation}
	\begin{aligned}
		\mathbf{B}_{\mathrm{l}}
		&=
		\mathrm{Conv}_{\mathrm{l}}^{1\times3}\big(\mathcal{P}_{\mathrm{l}}(\mathbf{U})\big),
		\quad
		\mathbf{B}_{\mathrm{r}}
		=
		\mathrm{Conv}_{\mathrm{r}}^{1\times3}\big(\mathcal{P}_{\mathrm{r}}(\mathbf{U})\big) \\
		\mathbf{B}_{\mathrm{t}}
		&=
		\mathrm{Conv}_{\mathrm{t}}^{3\times1}\big(\mathcal{P}_{\mathrm{t}}(\mathbf{U})\big),
		\quad
		\mathbf{B}_{\mathrm{b}}
		=
		\mathrm{Conv}_{\mathrm{b}}^{3\times1}\big(\mathcal{P}_{\mathrm{b}}(\mathbf{U})\big)
	\end{aligned}
	\label{eq:cdconv_branches}
\end{equation}
where $\mathbf{B}_{\mathrm{l}}$, $\mathbf{B}_{\mathrm{r}}$, $\mathbf{B}_{\mathrm{t}}$, and $\mathbf{B}_{\mathrm{b}}$ denote the outputs of the four directional branches, respectively. In Fig.~\ref{fig:cdconv_cdbr}, these operations correspond to branch-specific spatial convolutions (BSCs), each of which applies an asymmetric spatial convolution to a single directional branch. The output channels of the four branches are approximately evenly split and satisfy
\begin{equation}
	C_{\mathrm{l}}+C_{\mathrm{r}}+C_{\mathrm{t}}+C_{\mathrm{b}}=C_b
\end{equation}
The responses of the four directional branches are then concatenated along the channel dimension:
\begin{equation}
	\tilde{\mathbf{U}}
	=
	\operatorname{Cat}\big(
	\mathbf{B}_{\mathrm{l}},
	\mathbf{B}_{\mathrm{r}},
	\mathbf{B}_{\mathrm{t}},
	\mathbf{B}_{\mathrm{b}}
	\big)
	\label{eq:cdconv_cat}
\end{equation}
where $\operatorname{Cat}(\cdot)$ denotes channel-wise concatenation. The concatenated feature is first normalized and activated, and then aggregated by a lightweight channel fusion operation:
\begin{equation}
	\bar{\mathbf{U}}
	=
	\phi\big(\mathrm{BN}(\tilde{\mathbf{U}})\big)
	\label{eq:cdconv_bn}
\end{equation}
\begin{equation}
	\mathbf{U}_{\mathrm{cd}}
	=
	\phi\big(
	\mathrm{BN}(
	\mathrm{Conv}_{1\times1}(\bar{\mathbf{U}})
	)
	\big)
	\label{eq:channel_fuse}
\end{equation}
Note that Eq.~\eqref{eq:channel_fuse} corresponds to the Channel Fuse operation in Fig.~\ref{fig:cdconv_cdbr}. This process re-aggregates the responses of the four directional branches along the channel dimension to form a unified cross-shaped directional representation.

\begin{figure*}[!t]
	\centering
	\includegraphics[width=1.0\textwidth]{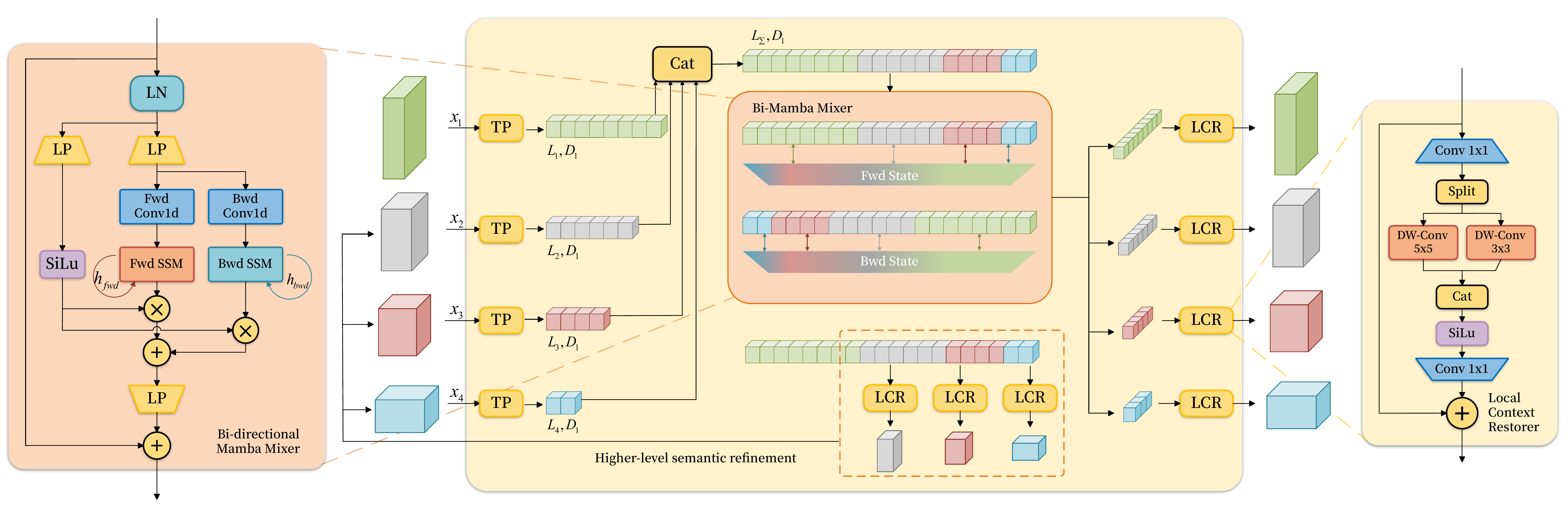}
	\caption{Detailed illustration of the proposed Latent Dense Cross-scale Fusion (LDCF) module. Multi-level features are projected into a unified latent space for Bi-Mamba-based dense cross-scale association, followed by hierarchy-aware semantic reorganization and local context restoration.}
	\label{fig:ldcf}
\end{figure*}

\subsubsection{Bottleneck Residual Design}

After direction-sensitive local enhancement, the block maps the feature back to the target channel dimension through another $1\times1$ convolution and fuses it with the shortcut branch via residual addition, yielding the final output:
\begin{equation}
	\mathbf{Y}
	=
	\phi\Big(
	\mathcal{F}_{\mathrm{exp}}(\mathbf{U}_{\mathrm{cd}})
	+
	\mathcal{S}(\mathbf{X})
	\Big)
	\label{eq:cdbr}
\end{equation}
where $\mathcal{F}_{\mathrm{exp}}(\cdot)$ denotes channel expansion, $\mathcal{S}(\cdot)$ denotes either the identity mapping or a $1\times1$ shortcut, and $\phi(\cdot)$ is the nonlinear activation function.

From a structural perspective, the version that keeps CDConv and residual connection but does not use channel compression and expansion is denoted as the cross-shaped directional residual block (CDResBlock). By further introducing bottleneck-style channel reduction and expansion, we obtain the proposed CDBR block. Compared with the non-bottleneck variant, this design confines direction-sensitive local enhancement to a controlled intermediate representation space, thereby reducing redundant feature expression while preserving local structure modeling ability. This principle is also consistent with residual bottlenecks and lightweight inverted residual structures in general vision, namely maintaining backbone expressiveness while controlling intermediate cost\cite{resnet,mobilenetv2}.

This design is especially important for IRSTD. First, the four-direction local modeling of CDConv better matches the local response distribution of weak tiny targets under complex backgrounds, thereby enhancing target preservation in shallow and middle-level features. Second, bottleneck compression constrains local enhancement to a more compact intermediate space, encouraging the network to allocate its limited representation capacity to discriminative structures relevant to target detection rather than redundant background textures. For scenes with extremely weak targets and heavy background interference, this combination of direction-sensitive local modeling and compact representation constraint is more conducive to forming stable weak-target representations.

\subsubsection{Hierarchical Encoder Construction}

Based on the proposed block, we construct a hierarchical encoder. The input image first passes through a shallow stem to establish initial local responses, and then multi-scale feature representations are progressively extracted through four stages of these blocks. As the feature hierarchy deepens, the network gains a larger receptive field while preserving and reinforcing weak-target local structure cues. The resulting multi-level features provide not only more robust inputs for subsequent latent cross-scale fusion, but also a stronger basis for detail restoration in the final decoding stage.

\subsection{Latent Dense Cross-scale Fusion}

Adequate cross-scale contextual dependency is indispensable for infrared small target detection, because weak tiny targets often require both local responses and surrounding context to be separated from cluttered backgrounds. However, continuously mixing multi-level features in a unified latent space may weaken their original scale attribution and make the subsequent decoder less stable. To address this issue, we propose a latent dense cross-scale fusion (LDCF) module, which establishes full-level contextual interaction while explicitly restoring hierarchy-aware multi-level semantics.

The structure of this module is summarized in Fig.~\ref{fig:ldcf} and consists of three successive steps. It first projects the deeper encoder features into a unified latent space and performs dense association with bidirectional Mamba modeling. It then reorganizes the interacted latent representation back into scale-specific feature maps, so that each branch keeps explicit hierarchical attribution after cross-scale interaction. Finally, high-level semantic branches are refined again to strengthen target-background discrimination before decoding. In this design, Mamba serves as a dedicated latent sequence mixer for higher-level cross-scale fusion rather than a backbone-wide context block.

\subsubsection{Latent Dense Association}

Let the multi-level features produced by the hierarchical encoder be
\begin{equation}
	\{\mathbf{x}_i\}_{i=1}^{4},
	\quad
	\mathbf{x}_i \in \mathbb{R}^{C_i \times H_i \times W_i}
\end{equation}
Following the shallow-feature routing strategy described above, the fusion module operates on the four deeper features, while $\mathbf{x}_0$ is preserved for final shallow restoration.

To establish cross-scale relations in a unified representation space, we first use a lightweight projection function $\mathcal{P}_i(\cdot)$ to map features at different levels into a shared latent space. In our implementation, Token Projection consists of a point-wise mapping branch and a depth-wise convolution branch, and is formulated as
\begin{equation}
	\mathbf{z}_i
	=
	\phi\Big(
	\mathrm{BN}\big(
	\mathrm{PW}_i(\mathbf{x}_i)
	+
	\mathrm{DW}_i\big(\mathrm{PW}_i(\mathbf{x}_i)\big)
	\big)
	\Big)
\end{equation}
where $\mathrm{PW}_i(\cdot)$ denotes a $1\times1$ point-wise projection, $\mathrm{DW}_i(\cdot)$ denotes a $3\times3$ depth-wise convolution, and $\phi(\cdot)$ denotes the SiLU activation. This yields
\begin{equation}
	\mathbf{z}_i \in \mathbb{R}^{D_1 \times H_i \times W_i}
\end{equation}
where $D_1$ is the shared latent dimension.
Next, the latent feature at each level is flattened and rearranged into a token sequence
\begin{equation}
	\mathbf{t}_i \in \mathbb{R}^{L_i \times D_1},
	\quad
	L_i = H_i W_i
\end{equation}
Since different feature levels have different spatial resolutions, the corresponding token sequence lengths $L_i$ also differ, while all sequences share the same latent dimension $D_1$. The token sequences are then concatenated along the length dimension to form a unified latent sequence
\begin{equation}
	\mathbf{T}
	=
	\operatorname{Cat}(\mathbf{t}_1,\mathbf{t}_2,\mathbf{t}_3,\mathbf{t}_4)
	\in
	\mathbb{R}^{L_{\Sigma} \times D_1},
	\quad
	L_{\Sigma} = \sum_{i=1}^{4} L_i
\end{equation}
To efficiently model long-range cross-scale dependencies in this unified latent space, we use a pre-normalized bidirectional Bi-Mamba mixer. This design is directly inspired by selective state-space modeling and its visual extensions\cite{mamba,visionmamba,vmamba}. We first apply LayerNorm to the input sequence:
\begin{equation}
	\tilde{\mathbf{T}} = \mathrm{LN}(\mathbf{T})
\end{equation}
We then perform state-space modeling in the forward and reverse directions, yielding the forward and backward hidden states
\begin{equation}
	\mathbf{h}_{\mathrm{fwd}}
	=
	\mathcal{M}_{\mathrm{fwd}}(\tilde{\mathbf{T}}),
	\quad
	\mathbf{h}_{\mathrm{bwd}}
	=
	\operatorname{Rev}\Big(
	\mathcal{M}_{\mathrm{bwd}}\big(
	\operatorname{Rev}(\tilde{\mathbf{T}})
	\big)
	\Big)
\end{equation}
where $\mathcal{M}_{\mathrm{fwd}}(\cdot)$ and $\mathcal{M}_{\mathrm{bwd}}(\cdot)$ denote the forward and backward Mamba transforms, respectively, and $\operatorname{Rev}(\cdot)$ denotes sequence reversal. The two state responses are averaged and incorporated with a scaled residual update to obtain the fused latent representation:
\begin{equation}
	\hat{\mathbf{T}}
	=
	\mathbf{T}
	+
	\gamma
	\cdot
	\frac{
		\mathbf{h}_{\mathrm{fwd}} + \mathbf{h}_{\mathrm{bwd}}
	}{2}
\end{equation}
where $\gamma$ is a learnable scaling parameter.

The key point of this process is that features from all scales are not modeled independently. Instead, they jointly participate in bidirectional state propagation within a shared latent sequence. As a result, cross-scale associations are not constructed through explicit dense links, but are implicitly established by forward and backward state updates in the unified latent space. In this way, shallow local responses and deep semantic context can interact within the same latent space, forming a more stable cross-scale discriminative representation.

\subsubsection{Hierarchy-Aware Semantic Reorganization}

After dense association in the unified latent space, the fused sequence representation is first split back into scale-specific latent sequences according to the original sequence length of each level:
\begin{equation}
	\{\hat{\mathbf{t}}_i\}_{i=1}^{4}
	=
	\operatorname{Split}(\hat{\mathbf{T}})
\end{equation}
These sequences are then reshaped into 2D latent feature maps according to the original spatial size of each level:
\begin{equation}
	\hat{\mathbf{z}}_i
	\in
	\mathbb{R}^{D_1 \times H_i \times W_i}
\end{equation}
On top of this, we further perform hierarchy-aware semantic reorganization and local restoration. This step is necessary because the preceding unified latent sequence strengthens cross-scale dependency but no longer explicitly preserves each branch as an independent feature level. Specifically, the fused latent representation is first restored to the corresponding 2D feature map of each level, then refined by a lightweight Local Context Restorer (LCR), and finally projected back to the original channel space with a residual link to the input branch:
\begin{equation}
	\mathbf{y}_i
	=
	\mathcal{O}_i\big(\mathcal{L}(\hat{\mathbf{z}}_i)\big)
	+
	\mathbf{x}_i
\end{equation}
where $\mathcal{L}(\cdot)$ denotes the LCR operation and $\mathcal{O}_i(\cdot)$ denotes the output projection. Here, LCR is used to recover local contextual consistency within each level, so that the reorganized features can preserve cross-scale associations while maintaining more stable local structural expression.

\subsubsection{High-Level Semantic Refinement}

Besides applying the above reorganization to all feature levels, we further perform grouped semantic refinement on higher-level latent representations. Specifically, after obtaining the initial reorganized features $\{\mathbf{y}_i\}_{i=1}^{4}$, we retain only the higher-level semantic branches $\{\mathbf{y}_i\}_{i=2}^{4}$ and project them again into a new shared latent space:
\begin{equation}
	\mathbf{z}_i'
	=
	\mathcal{P}_i'(\mathbf{y}_i),
	\quad
	i = 2,3,4
\end{equation}
\begin{equation}
	\mathbf{T}'
	=
	\operatorname{Cat}(\mathbf{t}_2',\mathbf{t}_3',\mathbf{t}_4')
	\in
	\mathbb{R}^{L_{\Sigma}' \times D_2}
\end{equation}
where $D_2$ denotes the latent dimension used in the high-level semantic refinement stage. The features are then modeled and reorganized again by a bidirectional Bi-Mamba mixer, yielding more stable high-level semantic representations. Therefore, the module does not stop at a single round of full-level interaction; instead, while preserving the shallowest local anchor, it further refines the higher-level semantic branches.

Overall, the multi-level outputs exhibit two desirable properties at the same time: they have already acquired sufficient cross-scale contextual dependency through the unified latent space, and they still preserve clear scale attribution and spatial structure. The core value of the fusion module therefore does not lie in merely adding one more cross-scale interaction stage. Rather, it lies in unifying dense association propagation in a shared latent space, hierarchy-aware semantic reorganization, and high-level semantic refinement, so that the fused features retain global associations while providing stable inputs for subsequent decoding and restoration.

\section{Experiments}

\subsection{Experimental Settings}
\subsubsection{Datasets}
We conduct experiments on three widely used public IRSTD datasets, namely IRSTD-1k~\cite{isnet}, NUAA-SIRST~\cite{acm}, and NUDT-SIRST~\cite{dnanet}. These datasets contain 1001, 427, and 1327 infrared images, respectively. Following the common protocol used in recent studies\cite{dnanet,hdnet,hstnet,mshnet}, IRSTD-1k is split into training and test sets with a ratio of 4:1, resulting in 800 training images and 201 test images. NUAA-SIRST uses the same split, with 341 training images and 86 test images, while NUDT-SIRST is evenly divided into training and test sets.

\begin{table*}[!t]
	\caption{Quantitative comparison with representative deep IRSTD methods on IRSTD-1k, NUAA-SIRST, and NUDT-SIRST. The dataset-level metric units are mIoU (\%), Pd (\%), and Fa ($10^{-6}$). Bold, underline, and ``--'' indicate the best result, the second-best result, and unavailable results, respectively.}
	\label{tab:sota_compare}
	\centering
	\footnotesize
	\setlength{\tabcolsep}{3.05pt}
	\renewcommand{\arraystretch}{1.15}
	\begin{tabular}{lccccccccccccccc}
		\toprule
		\multirow{2}{*}[-0.5ex]{Method} &
		\multirow{2}{*}[-0.5ex]{Pub.} &
		\multirow{2}{*}[-0.5ex]{Type} &
		\multirow{2}{*}[-0.5ex]{Params(M)$\downarrow$} &
		\multirow{2}{*}[-0.5ex]{FLOPs(G)$\downarrow$} &
		\multirow{2}{*}[-0.5ex]{Latency(ms)$\downarrow$}
		& \multicolumn{3}{c}{IRSTD-1k}
		& \multicolumn{3}{c}{NUAA-SIRST}
		& \multicolumn{3}{c}{NUDT-SIRST} \\
		\cmidrule(lr){7-9}\cmidrule(lr){10-12}\cmidrule(lr){13-15}
		& & & & &
		& mIoU$\uparrow$ & Pd$\uparrow$ & Fa$\downarrow$
		& mIoU$\uparrow$ & Pd$\uparrow$ & Fa$\downarrow$
		& mIoU$\uparrow$ & Pd$\uparrow$ & Fa$\downarrow$ \\
		\midrule
		ISNet~\cite{isnet}       & CVPR'22   & CNN      & \textbf{1.09}       & 122.55              & --                  & 61.85              & 90.24              & 31.56              & 70.49              & 95.06              & 67.98              & 81.24              & 97.78              & 6.34               \\
		DNANet~\cite{dnanet}      & TIP'22    & CNN      & 4.70                & 28.52               & 17.53               & 65.71              & 91.84              & 17.61              & 77.76              & 96.33              & 2.31               & 79.98              & 96.93              & 12.78              \\
		ISTDU-Net~\cite{istdunet}   & GRSL'22   & CNN      & 2.75                & 15.89               & 7.89                & 65.01              & \underline{93.94}  & 26.44              & 75.93              & 96.20              & 38.90              & 91.76              & 98.52              & 3.77               \\
		UIU-Net~\cite{uiunet}     & TIP'23    & CNN      & 50.54               & 108.85              & 10.13               & 68.69              & 91.25              & 13.48              & 77.53              & 92.40              & 9.33               & 75.91              & 96.83              & 18.61              \\
					MSHNet~\cite{mshnet}      & CVPR'24   & CNN      & 4.07                & 12.21               & \underline{7.57}    & 67.16              & 93.88              & 15.03              & 73.50              & 97.25              & 31.05              & 80.55              & 97.99              & 11.77              \\
		SCTransNet~\cite{sctransnet}  & TGRS'24   & CNN-T & 11.19               & 20.24               & 17.51               & 68.03              & 93.27              & 10.74              & 77.50              & 96.95              & 13.92              & 94.09              & 98.62              & 4.29               \\
		DATransNet~\cite{datransnet}  & GRSL'25   & CNN-T & 4.04                & \underline{10.90}   & 8.57                & 68.56              & 93.60              & 24.96              & --                 & --                 & --                 & 94.93              & 99.04              & \textbf{2.00}      \\
		HDNet~\cite{hdnet}       & TGRS'25   & CNN-D & 3.67                & 11.36               & 8.98                & \underline{70.26}  & \textbf{94.56}     & \textbf{4.33}      & 79.17              & \textbf{100.00}    & \textbf{0.53}      & 85.17              & 98.52              & 2.78               \\
		HSTNet~\cite{hstnet}      & TGRS'25   & CNN-T & --                  & --                  & --                  & 69.79              & 91.16              & 10.70              & 78.14              & \textbf{100.00}    & 14.72              & --                 & --                 & --                 \\
		FGARNet~\cite{fgar}       & TGRS'26   & CNN-D & 8.40                & 23.59               & 21.66               & 69.31              & 92.18              & \underline{4.75}   & \textbf{79.76}     & \textbf{100.00}    & 1.66               & 86.01              & 98.83              & 7.94               \\
					PQGNet~\cite{pqgnet}      & TGRS'26   & CNN-T & 1.19                & 19.78               & 9.43                & 67.51              & 91.50              & 28.70              & 76.60              & \underline{99.08}  & 5.68               & 95.27              & 98.31              & \underline{2.60}   \\
		FLINet~\cite{flinet}      & TGRS'26   & CNN   & 8.57                & 47.76               & 12.01               & 66.69              & 91.58              & 15.26              & 76.08              & 98.17              & 18.77              & \underline{95.52}  & \underline{99.15}  & 4.23               \\
		\midrule
		LCMamNet (Ours)        & --        & CNN-M    & \underline{1.18}    & \textbf{6.91}       & \textbf{6.62}       & \textbf{71.25}     & 93.20              & 11.31              & \underline{79.60}  & \textbf{100.00}    & \underline{0.71}   & \textbf{95.58}     & \textbf{99.47}     & 3.38               \\
		\bottomrule
	\end{tabular}
	
	\begin{minipage}{\textwidth}
		\vspace{4pt}
		\footnotesize
		\textit{Note:} CNN, CNN-T (Transformer), CNN-D (domain transform), and CNN-M (Mamba/SSM) denote different model types. Comparison results are obtained from public prediction maps, official implementations, or original reports when available. For remeasured complexity, the input size is $256\times256$, and FLOPs are counted as twice the number of multiply-accumulate operations (MACs). Latency denotes mean model forward inference latency measured with 50 warm-up runs and 200 timed runs. 
	\end{minipage}
\end{table*}

\subsubsection{Implementation Details}
The proposed method is implemented in PyTorch and trained/tested under Ubuntu 22.04 on a single NVIDIA RTX 4090 24GB GPU. All input images are resized to $256\times256$. During training, we apply random flipping and rotation for data augmentation.

The model is trained for 800 epochs with an early-stopping patience of 240 epochs. Both the training and test batch sizes are set to 12. We use AdamW as the optimizer, with an initial learning rate of $1\times10^{-3}$, a minimum learning rate of $1\times10^{-5}$, and a weight decay coefficient of $5\times10^{-2}$. A warmup plus cosine annealing schedule is adopted, where the warmup period lasts 5 epochs. The loss function is a weighted combination of BCE and Soft-IoU, with weights of 0.3 and 0.7, respectively.

\subsubsection{Evaluation Metrics}

We adopt mIoU, Pd, and Fa as the main quantitative metrics, and further analyze threshold-dependent detection behavior using ROC curves. Specifically, mIoU measures the pixel-level overlap between predictions and ground truth, Pd evaluates target-level detection ability, and Fa reflects the false-alarm level. Higher mIoU, Pd, and TPR, together with lower Fa and FPR, indicate better detection performance. For ROC analysis, TPR reflects target preservation and FPR reflects background leakage under varying decision thresholds, which are defined as
\begin{equation}
	\mathrm{TPR}(\tau)
	=
	\frac{\sum_{n=1}^{N}\mathrm{TP}_n^{(\tau)}}
	{\sum_{n=1}^{N}\left(\mathrm{TP}_n^{(\tau)}+\mathrm{FN}_n^{(\tau)}\right)}
\end{equation}
\begin{equation}
	\mathrm{FPR}(\tau)
	=
	\frac{\sum_{n=1}^{N}\mathrm{FP}_n^{(\tau)}}
	{\sum_{n=1}^{N}\left(\mathrm{TN}_n^{(\tau)}+\mathrm{FP}_n^{(\tau)}\right)}
\end{equation}
By traversing different thresholds $\tau$, we obtain ROC curves to compare the trade-off between target preservation and background suppression across methods.

In addition to accuracy metrics, we evaluate model efficiency and deployment potential using parameter count, GFLOPs, model forward inference latency, and inference speed on an edge platform.

\subsection{Comparison With State-of-the-Art Methods}
\subsubsection{Quantitative Results}
We first compare our method with representative deep IRSTD methods on IRSTD-1k, NUAA-SIRST, and NUDT-SIRST, as summarized in Table~\ref{tab:sota_compare}. The compared methods include CNN-based, Transformer-based, and domain-transform IRSTD models, as well as recent methods such as FGARNet~\cite{fgar}, PQGNet~\cite{pqgnet}, and FLINet~\cite{flinet}. Overall, our method remains among the strongest performers across the three datasets and achieves the most favorable accuracy-efficiency balance among the compared methods.

From the perspective of method categories, CNN-based models provide strong local representation and recovery paths, but their performance is still limited when small targets are dim or scale-varying. Transformer-based and hybrid methods further enhance global context modeling, yet they often introduce higher parameter counts, computation costs, or latency. By contrast, our method combines local structure encoding with latent cross-scale interaction, leading to stronger segmentation quality under a compact model budget.

A closer look at the three benchmarks further clarifies this trend. The advantage is most evident on IRSTD-1k, where weak targets and complex background interference make both missed detections and background leakage more likely. On this challenging dataset, our method achieves 71.25\% mIoU and outperforms strong baselines such as HDNet~\cite{hdnet}, HSTNet~\cite{hstnet}, FGARNet~\cite{fgar}, PQGNet~\cite{pqgnet}, and FLINet~\cite{flinet}. Although HDNet obtains a slightly higher Pd and lower Fa, our method achieves a higher mIoU, showing better region-level segmentation quality.

The remaining two benchmarks show that the improvement is not limited to one difficult case. NUAA-SIRST already yields very high Pd values for several advanced methods, making target-level detection close to saturation; under this condition, our method obtains a competitive mIoU of 79.60\% with the second-lowest Fa. NUDT-SIRST further validates the stability of target preservation on a larger benchmark, where our method obtains the best mIoU of 95.58\% and the best Pd of 99.47\%. Although a few methods report slightly lower Fa, the overall mIoU and Pd results show a more favorable balance between target completeness and background suppression.

Our method also maintains a favorable accuracy-efficiency balance. Compared with strong baselines such as SCTransNet~\cite{sctransnet}, HDNet~\cite{hdnet}, FGARNet~\cite{fgar}, PQGNet~\cite{pqgnet}, and FLINet~\cite{flinet}, it uses fewer parameters or lower computation while maintaining highly competitive segmentation accuracy. Specifically, the model requires only 1.18M parameters and 6.91 GFLOPs, corresponding to the second-smallest parameter count and the lowest reported FLOPs among methods with available complexity results. It also runs with the lowest measured mean model forward inference latency of 6.62 ms among the benchmarked methods. These results suggest that the performance gain is not obtained by simply enlarging the network, but by coordinating local structure encoding, latent cross-scale fusion, and selective shallow restoration within a compact framework.

\subsubsection{Visual Results and ROC Analysis}
\begin{figure*}[!t]
	\centering
	\includegraphics[width=\textwidth]{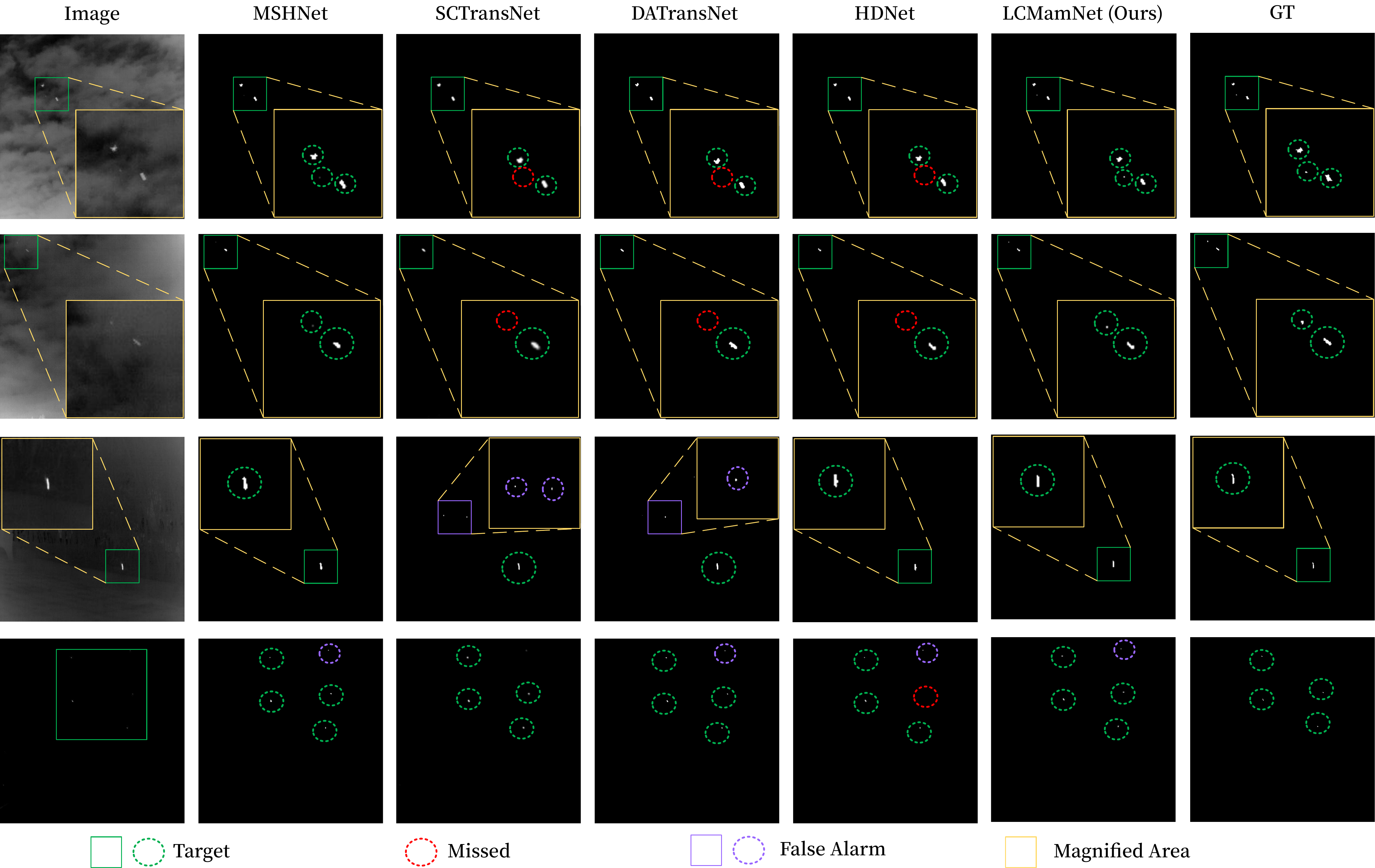}
	\caption{Qualitative comparison on IRSTD-1k. From left to right: Image, MSHNet~\cite{mshnet}, SCTransNet~\cite{sctransnet}, DATransNet~\cite{datransnet}, HDNet~\cite{hdnet}, LCMamNet (Ours), and GT.}
	\label{fig:qualitative}
\end{figure*}

Representative visual comparisons on IRSTD-1k are shown in Fig.~\ref{fig:qualitative}. The selected examples include cloud clutter, hazy background interference, and extremely dark scenes, where infrared small targets are easily confused with local bright noise or submerged by low-contrast backgrounds. These cases help reveal whether a method can preserve weak target responses without introducing excessive false alarms.

Compared with recent advanced methods, our method tends to preserve dim and small targets more completely and produce cleaner masks under complex backgrounds. In the examples with cloud interference, several competing methods tend to miss weak targets or generate incomplete responses, whereas our method keeps more continuous target regions. In the hazy and dark-background cases, it suppresses scattered background responses more effectively. These visual observations are consistent with the quantitative results, especially the higher mIoU on IRSTD-1k.

\begin{figure*}[!t]
	\centering
	\includegraphics[width=\textwidth]{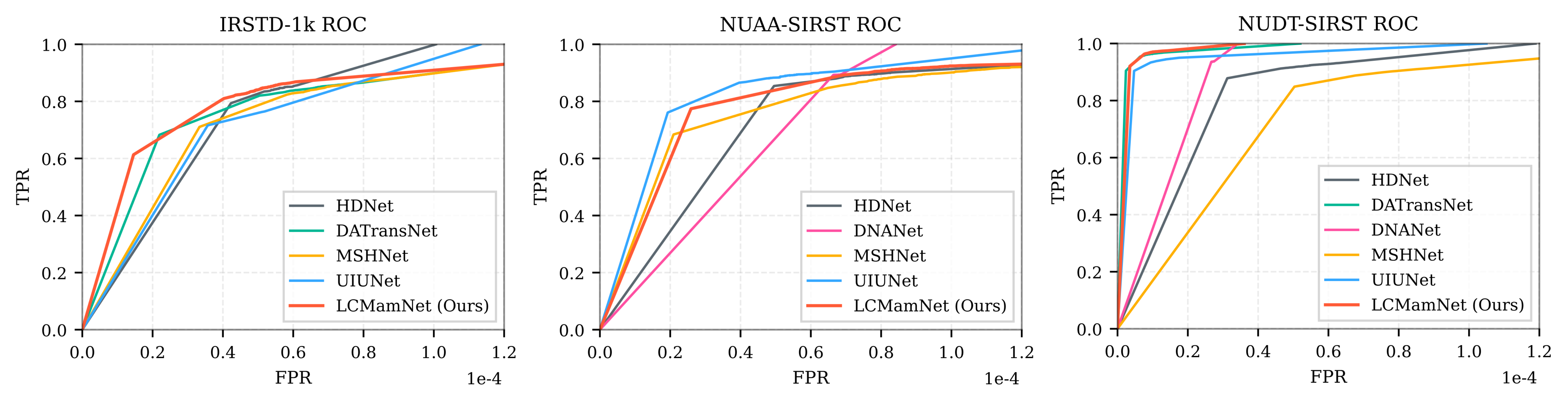}
	\caption{ROC curves on IRSTD-1k, NUAA-SIRST, and NUDT-SIRST.}
	\label{fig:roc_curves}
\end{figure*}

The threshold-dependent detection behavior is further compared on the three datasets in Fig.~\ref{fig:roc_curves}. Overall, our method shows competitive ROC curves and maintains a favorable balance between target preservation and background suppression, especially in the low-FPR region. This behavior indicates that the model does not rely on an overly aggressive foreground response to improve segmentation scores. Instead, it keeps a relatively stable trade-off under different thresholds, which is important for practical IRSTD applications where false alarms must be controlled.

\subsubsection{Efficiency and Edge Deployment}
Efficiency is another important factor for IRSTD models, especially for edge perception scenarios. In the comparison in Table~\ref{tab:sota_compare}, our method uses only 1.18M parameters and 6.91 GFLOPs, while achieving a mean model forward inference latency of 6.62 ms. Together with the lowest reported FLOPs, this latency advantage indicates that the lightweight design is effective not only in theoretical complexity but also in actual model execution.

We further evaluate practical inference speed on an NVIDIA Jetson Orin NX 16G SUPER under the MAXN power mode, with 50 model warm-up iterations followed by 200 measured iterations. The input shape follows batch size $\times$ channel $\times$ height $\times$ width. The deployment results in Table~\ref{tab:jetson_deploy} show that the PyTorch FP32 and FP16 implementations achieve comparable mIoU scores of 71.27\% and 71.18\%, respectively. After TensorRT FP16 deployment, the latency decreases to 9.75 ms, corresponding to 102.60 FPS, while the mIoU decreases to 65.96\%. This result indicates that TensorRT FP16 provides substantial acceleration but introduces a measurable accuracy drop, which may be related to reduced-precision inference and operator conversion for the CDBR encoder and Mamba-based latent fusion.

\begin{table}[!t]
	\caption{Deployment performance of our method on NVIDIA Jetson Orin NX 16G SUPER. mIoU is measured on IRSTD-1k.}
	\label{tab:jetson_deploy}
	\centering
	\scriptsize
	\setlength{\tabcolsep}{1.5pt}
	\begin{tabular*}{\columnwidth}{@{\extracolsep{\fill}}lccccc@{}}
		\toprule
		Backend & Precision & Input & mIoU$\uparrow$ & Latency(ms)$\downarrow$ & FPS$\uparrow$ \\
		\midrule
		PyTorch & FP32 & $1\times1\times256\times256$ & 71.27 & 49.86 & 20.06 \\
		PyTorch & FP16 & $1\times1\times256\times256$ & 71.18 & 47.05 & 21.25 \\
		TensorRT & FP16 & $1\times1\times256\times256$ & 65.96 & 9.75 & 102.60 \\
		\bottomrule
	\end{tabular*}
\end{table}

These deployment results demonstrate the practical potential of our method for real-time edge inference, while also showing that accuracy-preserving TensorRT optimization remains an important direction for resource-constrained IRSTD deployment.

\subsection{Ablation Studies}
We conduct ablation studies on IRSTD-1k to verify the effectiveness of the main design choices. This dataset contains weak targets and complex background interference, making it suitable for analyzing how different components affect target preservation and background suppression. The ablation studies are organized into four groups: encoder replacement, latent cross-scale fusion with semantic reorganization, token mixer replacement in LDCF, and full-framework synergy.

These experiments are designed to isolate different levels of contribution. The first group studies whether the proposed directional bottleneck encoder provides a better accuracy-efficiency basis. The second group examines whether latent dense association and hierarchy-aware reorganization are both necessary for effective cross-scale fusion. The third group further compares representative token mixers inside the same LDCF outer structure. The fourth group evaluates whether the final gain comes from the coordinated framework rather than a single isolated component.

\subsubsection{Effect of Encoder Replacement}

To isolate the effect of encoder design, Table~\ref{tab:ablation_encoder} compares three encoder units under the setting without latent cross-scale fusion and selective shallow restoration. ResBlock denotes a standard residual block, CDRes denotes a non-bottleneck residual block equipped with cross-shaped directional convolution, and CDBR denotes the final bottleneck version adopted in our model. Since the fusion module and the selective decoder are removed in this experiment, the comparison mainly reflects the standalone representation and complexity of different encoder units.

Compared with ResBlock, CDRes improves the mIoU from 66.01\% to 66.20\% while reducing the parameter count and GFLOPs to 2.14M and 12.21G, respectively. This result shows that directional local modeling is useful for preserving weak-target structures. The bottleneck version obtains a slightly lower isolated mIoU of 65.61\%, but further compresses the model to 1.45M parameters and 10.40G GFLOPs. Therefore, this table does not suggest that the bottleneck encoder is the strongest standalone block; rather, it shows that the block provides a compact and direction-aware encoder basis for the complete framework, where subsequent fusion and decoding can compensate for the mild compression cost.

\subsubsection{Effect of Latent Cross-scale Fusion and Semantic Reorganization}

The fusion module is evaluated under a fixed encoder and standard U-Net decoder in Table~\ref{tab:ablation_fusion}. Here, Assoc. denotes latent dense association, which performs bidirectional Mamba-based cross-scale interaction in the unified latent space. Direct denotes directly mapping the interacted latent features back to decoder features, while Reorg. denotes hierarchy-aware semantic reorganization before decoding.

\begin{table}[!t]
	\caption{Ablation on latent cross-scale fusion and semantic reorganization on IRSTD-1k.}
	\label{tab:ablation_fusion}
	\centering
	\scriptsize
	\begin{tabular*}{\columnwidth}{@{\extracolsep{\fill}}lcccccc@{}}
		\toprule
		Encoder & Assoc. & Bridge & Decoder & mIoU$\uparrow$ & \makecell{Params(M)$\downarrow$} & \makecell{FLOPs(G)$\downarrow$} \\
		\midrule
		CDBR & --  & --     & U\text{-}Net & 65.61 & \textbf{1.45} & \textbf{10.40} \\
		CDBR & Yes & --     & U\text{-}Net & 67.36 & 1.65 & 12.03 \\
		CDBR & Yes & Direct & U\text{-}Net & 67.70 & 1.59 & 11.84 \\
		CDBR & Yes & Reorg. & U\text{-}Net & \textbf{68.59} & 1.73 & 12.26 \\
		\bottomrule
	\end{tabular*}
\end{table}

\begin{table}[!t]
	\caption{Ablation on encoder replacement on IRSTD-1k. LDCF and SBA are removed to isolate the encoder effect.}
	\label{tab:ablation_encoder}
	\centering
	\scriptsize
	\begin{tabular*}{\columnwidth}{@{\extracolsep{\fill}}lccc@{}}
		\toprule
		Encoder & mIoU$\uparrow$ & \makecell{Params(M)$\downarrow$} & \makecell{FLOPs(G)$\downarrow$} \\
		\midrule
		ResBlock & 66.01 & 3.39 & 15.86 \\
		CDRes    & \textbf{66.20} & 2.14 & 12.21 \\
		CDBR     & 65.61 & \textbf{1.45} & \textbf{10.40} \\
		\bottomrule
	\end{tabular*}
\end{table}

Without latent cross-scale fusion, the model obtains an mIoU of 65.61\%. Introducing latent dense association raises the mIoU to 67.36\%, indicating that cross-scale context helps separate weak targets from complex backgrounds. Passing the interacted features through a direct bridge further improves the mIoU to 67.70\%, which shows that the interacted latent representation can already provide useful contextual information for decoding.

However, direct bridging does not explicitly recover the semantic hierarchy among different feature levels after latent interaction. Replacing it with semantic reorganization increases the mIoU to 68.59\%, confirming that cross-scale association alone is not sufficient. The fused representation also needs to be reorganized into scale-aware hierarchical features before being sent to the decoder. This progressive gain supports the design of combining latent interaction with hierarchy-aware reorganization.

\subsubsection{Effect of Token Mixer in LDCF}

The token mixer in LDCF needs to associate a long multi-resolution latent sequence while keeping the fusion module compact. Under the same LDCF outer structure and training protocol, Table~\ref{tab:ablation_mixer} compares representative mixer choices for this sequence association step. Efficient Self-Attn denotes a channel self-attention mixer that forms attention along the channel dimension to avoid full token-token attention over the concatenated latent sequence, and Bi-GRU replaces state-space modeling with a standard bidirectional GRU mixer~\cite{cho_gru}.

\begin{table}[!t]
	\caption{Ablation on token mixers in LDCF on IRSTD-1k.}
	\label{tab:ablation_mixer}
	\centering
	\scriptsize
	\begin{tabular*}{\columnwidth}{@{\extracolsep{\fill}}lcccc@{}}
		\toprule
		Mixer & Type & mIoU$\uparrow$ & \makecell{Params(M)$\downarrow$} & \makecell{FLOPs(G)$\downarrow$} \\
		\midrule
		Efficient Self-Attn & Channel self-attn. & 69.47 & \textbf{1.09} & 7.88 \\
		Bi-GRU & Bi. recurrent & 70.47 & 1.15 & 9.92 \\
		Bi-Mamba & Bi. SSM & \textbf{71.25} & 1.18 & \textbf{6.91} \\
		\bottomrule
	\end{tabular*}
\end{table}

Bi-Mamba achieves the highest mIoU of 71.25\% with 1.18M parameters and 6.91 GFLOPs. Efficient Self-Attn has fewer parameters, but its mIoU is lower and its FLOPs are higher than Bi-Mamba. Bi-GRU obtains a closer mIoU of 70.47\%, but it increases the computation to 9.92 GFLOPs. This gap is consistent with the sequence property of LDCF: bidirectional Mamba models the long latent sequence through linear state-space scanning, whereas gated recurrent units process the flattened sequence with heavier step-wise recurrence. These results support the use of Bi-Mamba as the token mixer in LDCF, since it provides the best accuracy-complexity trade-off among the compared alternatives.

\subsubsection{Synergy Analysis in the Full Framework}

The previous groups isolate encoder, fusion, and mixer behavior under controlled settings. To further verify whether these designs remain effective in the final network, we perform a controlled comparison under the full framework in Table~\ref{tab:ablation_synergy}. In this table, the row with CDBR, Reorg., and U-Net removes the selective decoder, the row with CDRes tests whether the non-bottleneck encoder is still preferable in the full framework, and the row with Direct replaces semantic reorganization with direct bridging.

\begin{table}[!t]
	\caption{Ablation on full-framework synergy on IRSTD-1k. SBA denotes synergistic bottleneck attention.}
	\label{tab:ablation_synergy}
	\centering
	\scriptsize
	\begin{tabular*}{\columnwidth}{@{\extracolsep{\fill}}lccccc@{}}
		\toprule
		Encoder & Bridge & Decoder & mIoU$\uparrow$ & \makecell{Params(M)$\downarrow$} & \makecell{FLOPs(G)$\downarrow$} \\
		\midrule
		CDBR  & Reorg. & U\text{-}Net & 68.59 & 1.73 & 12.26 \\
		CDRes & Reorg. & SBA          & 70.13 & 1.86 & 8.72 \\
		CDBR  & Direct & SBA          & 68.71 & \textbf{0.93} & \textbf{6.48} \\
		CDBR  & Reorg. & SBA          & \textbf{71.25} & 1.18 & 6.91 \\
		\bottomrule
	\end{tabular*}
\end{table}

The full model achieves 71.25\% mIoU with 1.18M parameters and 6.91 GFLOPs. Replacing the bottleneck encoder with the non-bottleneck CDRes block reduces the mIoU to 70.13\% and increases the parameter count to 1.86M, showing that the compact encoder is more suitable once latent fusion and selective decoding are introduced. This result also explains why the isolated encoder comparison in Table~\ref{tab:ablation_encoder} should be interpreted from an accuracy-efficiency perspective rather than by standalone mIoU alone.

The remaining variants further validate the two downstream designs. Replacing semantic reorganization with a direct bridge decreases the mIoU to 68.71\%, confirming that restoring hierarchical semantics is still important in the full framework. Replacing the selective decoder with a standard U-Net decoder reduces the mIoU to 68.59\%, indicating that the proposed shallow-feature routing and skip regulation are necessary for restoring useful details without introducing excessive background clutter. Overall, these results verify that the final advantage comes from the synergy among local structure encoding, latent cross-scale fusion, and selective decoding.

\section{Conclusion}
This article proposes LCMamNet, a lightweight network for infrared small target detection that aims to improve weak-target preservation, cross-scale context modeling, and background suppression under an efficient inference budget. Instead of relying on a single stronger module, the method coordinates direction-aware local structure encoding, latent cross-scale interaction, and selective shallow feature restoration in a unified encoder-fusion-decoder framework. Experiments on IRSTD-1k, NUAA-SIRST, and NUDT-SIRST show that our method achieves highly competitive segmentation performance while maintaining low parameters, computation, and measured inference latency. The ablation studies further verify that the final performance gain comes from the complementary effects of the compact encoder, the latent fusion module, and the selective decoding path. Deployment results on an NVIDIA Jetson Orin NX 16G SUPER also demonstrate its practical potential for real-time edge inference. Despite these results, the current training and evaluation mainly follow segmentation-oriented objectives and metrics. For very small targets with blurred edges, the predicted mask may still not perfectly match the detailed shape of the original infrared target. Future work will explore shape-aware metrics and architecture designs to improve contour fidelity for such tiny targets.

\bibliographystyle{IEEEtran}
\bibliography{references}
\end{document}